\documentclass[journal]{IEEEtran}
\usepackage{cite}

\usepackage{multirow}

\usepackage{subfigure}

\usepackage{epstopdf}

\ifCLASSINFOpdf
   \usepackage[pdftex]{graphicx}
\else
	\usepackage[caption=false,font=footnotesize]{subfig}
\fi
\usepackage{amsfonts}
\usepackage{amsmath}
\ifCLASSOPTIONcompsoc
  \usepackage[caption=false,font=normalsize,labelfont=sf,textfont=sf]{subfig}
\else
  \usepackage[caption=false,font=footnotesize]{subfig}
\fi
\usepackage{url}

\begin{document}
%
\title{SAN: Scale-Aware Network for Semantic Segmentation of High-Resolution Aerial Images}
%
%
%

\author{	
	\IEEEauthorblockN{
		Jingbo Lin\IEEEauthorrefmark{1},
		Weipeng Jing\IEEEauthorrefmark{2},~\IEEEmembership{Member,~IEEE}, and 
		Houbing Song\IEEEauthorrefmark{3},~\IEEEmembership{Senior,~IEEE}
	}
	
	\IEEEauthorblockA{\IEEEauthorrefmark{1}\IEEEauthorrefmark{2}College of Information and Computer Engineering, Northeast Forestry University, Harbin, HLJ China}
	
	\IEEEauthorblockA{\IEEEauthorrefmark{3}Department of Electrical, Computer, Software, and Systems Engineering, Embry-Riddle Aeronautical University, Daytona Beach, FL 32114 USA}
	
\thanks{M. Shell was with the Department
of Electrical and Computer Engineering, Georgia Institute of Technology, Atlanta,
GA, 30332 USA e-mail: (see http://www.michaelshell.org/contact.html).}
\thanks{J. Doe and J. Doe are with Anonymous University.}
\thanks{Manuscript received April 19, 2005; revised August 26, 2015.}}

%
%

\markboth{IEEE GEOSCIENCE AND REMOTE SENSING LETTERS}%
{Shell \MakeLowercase{\textit{et al.}}: Bare Demo of IEEEtran.cls for IEEE Journals}
%



\maketitle

\begin{abstract}
High-resolution aerial images have a wide range of applications, such as military exploration, and urban planning.
Semantic segmentation is a fundamental method extensively used in the analysis of high-resolution aerial images. However, the ground objects in high-resolution aerial images have the characteristics of inconsistent scale, and this feature usually leads to unexpected predictions.
To tackle this issue, we propose a novel scale-aware module (SAM).
In SAM, 
we employ re-sampling method aimed to make pixels adjust their positions to fit the ground objects with different scales, and it implicitly introduce spatial attention by employing re-sampling map as weighted map.
As a result, the network with the proposed module named scale-aware network (SANet) has a stronger ability to distinguish the ground objects with inconsistent scale. Other than this, our proposed modules can easily embed in most of the existing network to improve their performance.
We evaluate our modules on International Society for Photogrammetry and Remote Sensing Vaihingen Dataset, and the experimental results and comprehensive analysis demonstrate the effectiveness of our proposed module. 

\end{abstract}


%
\IEEEpeerreviewmaketitle

\section{Introduction}
%
%
%
%
\IEEEPARstart{S}{emantic} segmentation has great significance in high-resolution remote sensing images. 
As one of the most critical methods of interpretation and analysis of remote sensing images, it was widely used in environmental monitoring and smart cities.
With the development of remote sensing technologies, advanced sensors can provide more and more high-resolution and high-quality images. 
Due to the limitation of expressing ability, the conventional segmentation method that over depends on handcraft feature can no longer work well with large-scale remote sensing data and the complex appearance variations of ground objects.

In recent years, the deep convolutional neural networks (DCNN) have shown their outstanding performance in many vision tasks, such as image classification\cite{Krizhevsky2012}, object detection\cite{lin_microsoft_2014}, and semantic segmentation\cite{everingham_pascal_2010}.
From the milestone work of Long \emph{et al.}\cite{long_fully_nodate} in 2014, the fully convolutional neural networks (FCNs) have been extensively employed in semantic segmentation tasks. Inspired by Ronneberger \emph{et al.}\cite{ronneberger_u-net:_2015}, the architecture of encoder-decoder is widely used and has varieties of variants, such as SegNet\cite{badrinarayanan_segnet:_2015} and DeconvNet\cite{noh_learning_2015}. 
To get better predictions, the DeepLab family utilize dense conditional random field (dense-CRF)\cite{chen_deeplab:_2016} as post-processing and propose atrous spatial pyramid pooling (ASPP) module\cite{chen_encoder-decoder_2018} to aggregate multi-scale feature better.
RefineNet\cite{lin_refinenet:_2016} and GCN\cite{peng_large_2017} adopt multi-stage architecture to refine the confidence map stage by stage.

Although the methods of previous works can get more and more accurate prediction in scene parsing task, there still exists challenging issues in semantic segmentation of high-resolution remote sensing images.
Different from the typical indoor scenes and outdoor scenes, the high-resolution remote sensing images have complex ground objects with the characteristics of inconsistent scale.
For the problem of large margin difference in scales, the lack of receptive field causes incompletely identified of large-scale objects; 
the receptive fields far larger than the small-scale objects introduce too much irrelevant information, and it usually leads to small-scale objects unrecognized.
Therefore, the networks used for segmentation in high-resolution remote sensing images should have the scale-aware ability to objects with different scales.
\begin{figure*}[!t]
	\centering
	\includegraphics[]{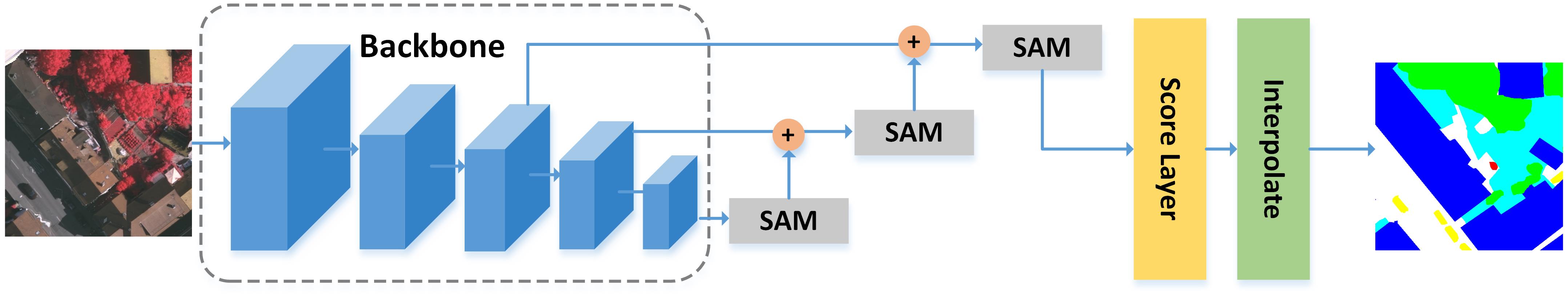}
	\caption{Overall architecture of the proposed SANet.}
	\label{fig_overall}
\end{figure*}
\begin{figure}[!t]
	\centering
	\includegraphics[]{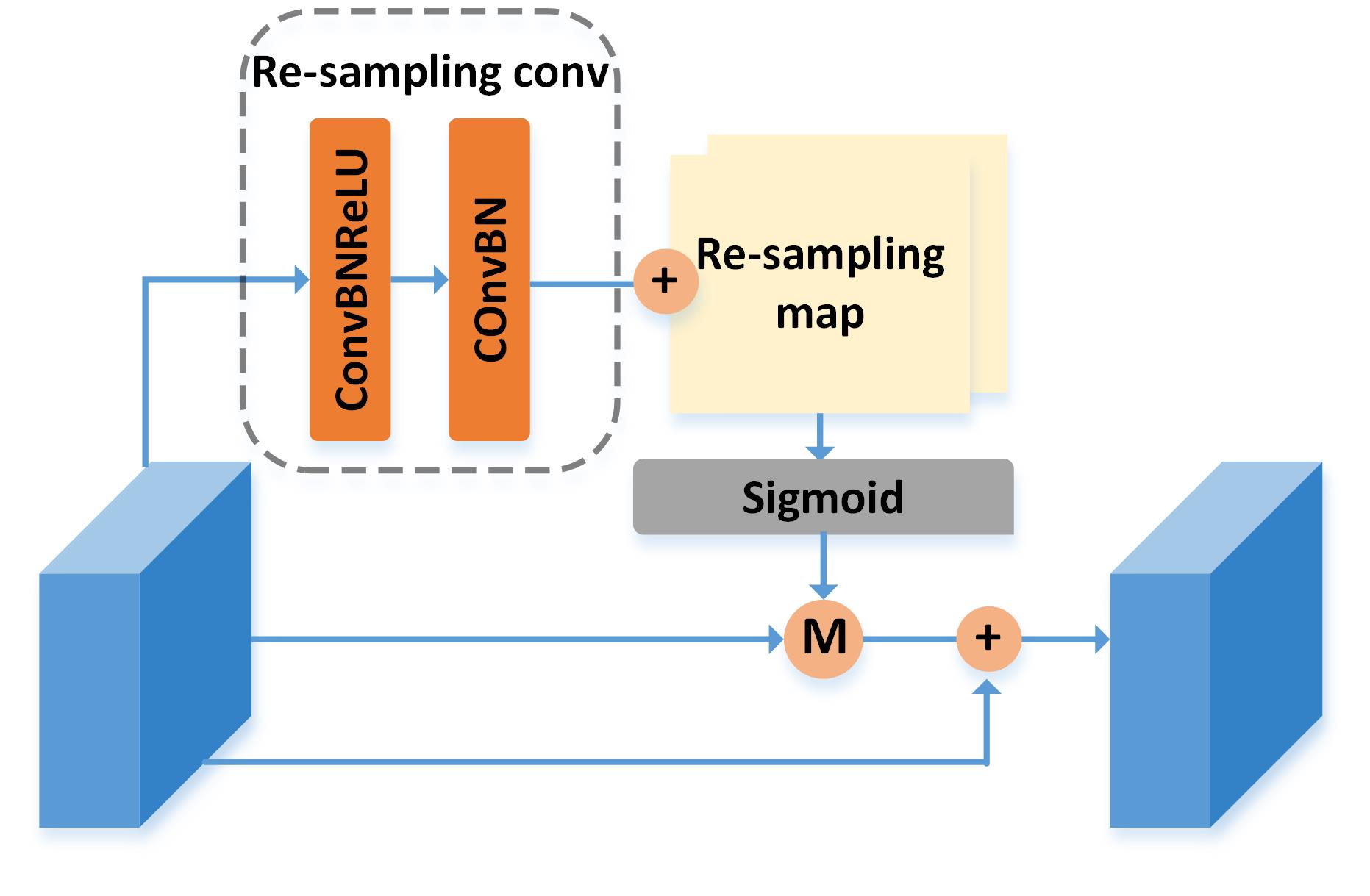}\label{fig_sam}
	\hfill
	\caption{The structure of the proposed adaptive Scale-Aware Module (SAM). The circles with 'M' means element-wise multiplication.}
\end{figure}

It is difficult to obtain appropriate information with only single scale features because of the existence of objects in multiple scales. 
To tackle this issue, the conventional approaches adopt weighted summation to aggregate multi-scale information . 
Nogueira \emph{et al.}\cite{nogueira_dynamic_2019} proposed a method that fuse multi-scale information by weighted summation of distinct size of patches. 
Li \emph{et al.}\cite{li_adaptive_2019} also adopted the weighted summation method in their adaptive multi-scale deep fusion module.
Lu \emph{et al.}\cite{lu_feature_2019} proposed supervised strategy, utilizing the semantic label information to aggregation information progressively.
Different from the mentioned methods, we introduce scale-aware ability to existing networks.
There are some similar works for scene parsing tasks in recent years.
Jaderberg \emph{et al.}\cite{jaderberg_spatial_2015} proposed a learnable module \emph{Spatial Transformer} that allows the spatial manipulation of pixels within feature maps by learning affine transformation.
Based on the Jaderberg’s work, Wei \emph{et al.}\cite{wei_learning_2017} use two affine transformation layers to regulate receptive field automatically.
But the method based on affine transformation is more suitable for the objects with different rotation transformation.
In Zhang’s work\cite{zhang_scale-adaptive_2017}, they introduce scale-adaptive mechanism by additional scale regression layers that can dynamically infer the position-adaptive scale coefficients to shrink or expand the convolutional patches. 
However, the sampling space has limitation since the scale coefficient treats the horizontal and vertical axes equally in 2D space.

In this letter, we propose a module called SAM, which can be trained end-to-end and easily embedded in existing networks.
Different from the previous works, our SAM expands the sampling space by learning two-dimensional re-sampling maps that consider horizontal and vertical axes differently. 
In SAM, every pixels are re-sampled based on the re-sampling maps, which is learned from network and input data. That means, our SAM not only has position-adaptive ability to fit objects with various sizes but also has data-adaptive ability to adjust different testing images.
We employ scale-aware module in FCN and conduct experiment on ISPRS Vaihingen Dataset. Experimental results show the effectiveness of our proposed scale-aware module SAM.

\section{Proposed Methods}
\subsection{Overview}
Without bells and whistles, the overall architecture of our SANet is shown in Fig. \ref{fig_overall}.
The backbone as a feature extractor, it can be any of networks used for classification.
Following with FCN8s, the downsampling rate is 32, and we employ scale-aware module at the end of each stage. At last, we use a simple score layer and upsampling layer to get the final prediction.
\subsection{Scale-Aware Module}
In this section, we will elaborate the implementation details and feasibility of end-to-end training of our proposed module.
The structure of SAM is shown in Fig. \ref{fig_sam}.
It generally takes two steps to get the scale-aware feature maps.

The first step is learning the re-sampling map by re-sampling convolutions.
Supposing we take $\mathbf{I}\in\mathbb{R}^{H\times W\times C}$ as input, where $H$ and $W$ is spatial size of feature map, $C$ is the number of input channel.
Going through re-sampling convolutions, we will get the re-sampling information $\mathbf{S}\in\mathbb{R}^{H\times W\times 2}$ for each pixel in horizontal and vertical axes, and then merging re-sampling information into re-sampling map by element-wise addition.
It is worth noting that, we initial the re-sampling map by mapping the coordinates of each pixel in original feature maps to range of [-1, 1]. That is, the coordinate of top-left position is (-1, -1) and the bottom-right position is (1, 1).
In order to guarantee pixels tune their spatial position based on the original and avoid excessive offset,
we employ normal distribution $\mathcal{N}$ with small std as initialization strategy of the re-sampling convolutional layers, that can be formulated as:
\begin{equation}
\omega\sim \mathcal{N}(0,\sigma^2), \sigma \ll 1
\end{equation}
where $\omega$ is weights of convolutional layers. We take $\sigma$ as 0.001 and ignore the bias term in our experiments. Under this initialization scheme, the generated re-sampling information is regulated in [-1, 1]. Thus, the re-sampled pixels move start from original position gradually during training. Moreover, we also perform clamp operation during training to avoid the unexpected conditions.

The second step is aimed to re-sample pixels from the original feature maps based on the re-sampling map. Inspired by the work of Jaderberg \emph{et al.}, we also adopt bilinear interpolation as our sampling method. Considering the kernel $\mathbf{K}\in\mathbb{R}^{k_h\times k_w\times C_k}$ and the patches used to convolve $\mathbf{P}\in\mathbb{R}^{k_h\times k_w\times C_p}$, where $k_h$ and $k_w$ is the spatial size of filters, $C_k$ and $C_p$ is the number of channels of filter and feature maps, respectively. The standard convolution without the SAM-transformed feature maps can be expressed as:
\begin{equation}
	C(K, P) = \sum_{ck,cp}\sum_{i,j} (k^{ij,ck}\times p^{ij,cp} + b^{ck})
	\label{eq_2}
\end{equation}
where $k^{ij,ck}$ and $p^{ij,cp}$ are the pixel at the position (i, j) of $ck$, $cp$ channel in filters and feature maps, respectively.
For convenient, we only consider the spatial dimension and ignore the bias term. Mapping the spatial positions of pixels in original feature maps into [-1, 1], the coordinates of pixel $p$$(p_x, p_y)$ in re-sampled feature maps are:
\begin{equation}	
\left\{
		\begin{array}{lr}
			x^{ij} = p_x + s_x	\\
			y^{ij} = p_y + s_y   &
		\end{array}
\right.
\end{equation}
where $(p_x, p_y)$ is the mapped coordinate of the original position of pixel $p$, $s_x$ and $s_y$ is the corresponding re-sampling information learned by re-sampling convolutions.
Then we perform bilinear interpolation to get SAM-transformed pixels based on re-sampling map, the value of transformed pixels can be calculated as:
\begin{equation}
	V(p^{ij}) = \sum_{q}B(p^{ij}, q)\cdot v(q)	
	\label{eq_4}
\end{equation}
\begin{equation}
	B(p^{ij},q) = b(x^{ij}, q_x)\cdot b(y^{ij}, q_y)
	\label{eq_5}
\end{equation}
where $B(\cdot)$ denotes bilinear interpolation, $b(p,q)=max(0,1-|p-q|)$ and $v(q)$ is the value of pixels $q$ that participant interpolation.
Instead of directly using the re-sampled feature maps to participate the following calculation, we employ residual learning scheme.
The re-sampled feature maps are converted to weight maps by sigmoid layer and element-wise multiplication with the original feature maps.
On one hand, it maintains the original information and better merges with the re-sampled information. On the other hand, the residual learning is benefit for optimizing. Finally, the convolution with the SAM-transformed feature maps can be expressed as:
\begin{equation}
\left\{\begin{array}{lr}
		C(K,P) = C(K, T(P)) \\
		T(P) = v(P) + v(P) \times \sigma( V(M(P)))& 
		\end{array}	
\right.
\label{eq_6}
\end{equation}
where $C(K, P)$ is followed with Eq.\ref{eq_2}, $T(P)$ represents the SAM-transformed pixels, and $M(\cdot)$ as mapping.
To elaborate the differentiable of our proposed method, we denote $K_{ij}$ as the corresponding matrix of kernels \textbf{K} in spatial-wise, and $P_{ij}$ as the related patches. For standard convolution, the forward propagation is:
\begin{equation}
	O = C(K_{ij}, P_{ij}) = \sum_{i,j} K_{ij} \cdot P_{ij} + b
\end{equation} 
During backward propagation, given the gradient of output $O$, the gradient of each value can be obtained (here we also ignore the bias):
\begin{equation}
\left\{\begin{array}{lr}
		g(P_{ij}) = \sum (K_{ij})^T\cdot g(O) \\
     	g(K_{ij}) = \sum g(O)\cdot (P_{ij})^T	& \\
		\end{array}
\right.
\end{equation}
where $g(\cdot)$ denotes gradient function, and $(\cdot)^T$ is the matrix transposition. According to Eq. \ref{eq_2} and \ref{eq_6}, the gradient of each value in convolution of SAM-transformed feature map is:
\begin{equation}
\left\{\begin{array}{lr}
		g(P_{ij}) = \sum (K_{ij})^T\cdot	g(T(P_{ij}))\cdot g(O) \\
		g(K_{ij}) = \sum g(O)\cdot T(P_{ij})^T &\\
	\end{array}
\right.
\end{equation}
then we denote $Z = \sigma(V(M(\cdot)))$, and $g(T(P_{ij}))$ can be expressed as:
\begin{equation}
	g(T(P_{ij})) = 1 + Z(P_{ij}) + P_{ij}\cdot g(Z(P_{ij}))
\end{equation}
we know sigmoid is differentiable and following Eq.\ref{eq_5} the partial derivatives of input pixels used for interpolation are:
\begin{equation}
	\frac{\partial Z_{ij}}{\partial Q_{ij}} = max(0,1-|x^{ij}-q_x|)max(0,1-|y^{ij}-q_y|)
\end{equation}
and the partial derivatives w.r.t. coordinates of each re-sampled pixel $p$:
\begin{equation}
	\frac{\partial Z_{ij}}{\partial x^{ij}} = \sum Q_{ij}max(0,1-|y^{ij}-q_y|) g(B_x(x^{ij},q_x)) 
\end{equation}
\begin{equation}
g(B_x(x^{ij}, q_x)) =\left\{
	\begin{array}{lr}
	0, & |x^{ij}-q_x|\geq 1 \\
	1, & x^{ij} > q_x, |x^{ij}-q_x|< 1\\
	-1,& x^{ij} < q_x, |x^{ij}-q_x|< 1
	\end{array}	
\right. 
\end{equation}
Furthermore, the coordinates of the re-sampled pixels are generated by the original coordinates and the re-sampling information, which is learned by the re-sampling convolutions. The gradient of input feature maps $g(P_{ij})$ is:
\begin{equation}
g(P_{ij})=\sum (K_{ij})^T(1+Z(P_{ij})+P_{ij}\cdot g(Z_{ij}))\cdot g(O)
\label{eq_14}
\end{equation}

As the Eq. \ref{eq_14} shown that, our proposed method can be trained end-to-end, optimized by stochastic gradient descent algorithms, and does not need any additional supervisions. Other than this, the residual-block-like structure make gradient flow easily. We employ re-sampling map as weighted map let SAM implicitly introduce spatial-wise attention mechanism. From this perspective, it also makes sense to explain the scale-aware ability of our proposed module.

More importantly, SAM make networks have scale adaptive ability. In conventional methods that applied DCNNs, the parameters are learned from training data, and they are fixed in testing phase. But in SAM, the re-sampling map is not directly learned from training data, it is generated by the input and the network learned from training data.
Therefore, the network with SAM not only has better scale-aware ability but also is self-adaptive to different testing images.

\begin{table*}[!t]
	\renewcommand{\arraystretch}{1.3}
	\caption{The Experiment Results of Ablation Studies, Per-Class IoU(\%), Per-Class F1-Score(\%), Mean IoU(\%), Mean F1-Score(\%), and Overall Accuracy(\%).}
	\label{table_ablations}
	\centering
	\begin{tabular}{l c c c c c c c c c c c c c}
		\hline\hline
		\multirow{2}*{Networks}&\multicolumn{2}{c}{Imp surf}&\multicolumn{2}{c}{Building}&\multicolumn{2}{c}{Low Veg}&\multicolumn{2}{c}{Tree}&\multicolumn{2}{c}{Car}&\multicolumn{2}{c}{Avg.} &\multirow{2}*{Overall Acc} \\
		                             
		&IoU &F1&IoU &F1 &IoU &F1 &IoU &F1 &IoU &F1 &mean IoU &mean F1 &\\
		\hline
		FCN8s         & 73.07 & 84.29 & 80.23 & 88.90 & 56.02 & 71.46 & 70.47 & 82.42 & 51.10 & 67.32 & 66.18 & 78.88 & 83.76\\
		FCN8s-SAM-SC  & 73.13 & 84.31 & 80.93 & 89.33 & 56.22 & 71.61 & 70.63 & 82.53 & 53.40 & 69.24 & 66.86 & 79.41 & 83.98 \\
		FCN8s-SAM-S   & 73.21 & 84.39 & 80.77 & 89.23 & 56.47 & 71.79 & 70.77 & 82.64 & 51.17 & 67.33 & 66.48 & 79.08 & 83.94\\
		FCN8s-SAM-MC  & 73.25 & 84.40 & 80.78 & 89.23 & 57.11 & 72.34 & 71.50 & 83.20 & 51.64 & 67.68 & 66.86 & 79.37 & 84.17\\
		FCN8s-SAM-M   & 75.28 & 85.77 & 82.25 & 90.13 & 58.81 & 73.75 & 72.46 & 83.82 & 53.97 & 69.76 & 68.55 & 80.65 & 85.12\\	

		\hline\hline
	\end{tabular}
\end{table*}
\begin{table*}[!t]
	\renewcommand{\arraystretch}{1.3}
	\caption{The Comparison With The State-Of-The-Art Networks, Per-Class IoU, Per-Class F1-Score, Mean IoU, Mean F1-Score, and Overall Accuracy.}
	\label{table_evalation}
	\centering
	\begin{tabular}{l c c c c c c c c c c c c c}
		\hline\hline
		\multirow{2}*{Networks}&\multicolumn{2}{c}{Imp surf}&\multicolumn{2}{c}{Building}&\multicolumn{2}{c}{Low Veg}&\multicolumn{2}{c}{Tree}&\multicolumn{2}{c}{Car}&\multicolumn{2}{c}{Avg.} &\multirow{2}*{Overall Acc}\\
		
		&IoU &F1&IoU &F1 &IoU &F1 &IoU &F1 &IoU &F1 &mean IoU &mean F1 & \\
		\hline
		LWRefineNet   & 65.83 & 79.22 & 73.69 & 84.70 & 47.84 & 64.21 & 63.28 & 77.20 & 34.88 & 51.16 & 57.10 & 71.30 & 78.52\\
		DeepLabv3     & 73.23 & 84.40 & 79.80 & 88.64 & 54.34 & 70.02 & 69.32 & 81.64 & 50.58 & 66.69 & 65.45 & 78.28 & 83.22\\  
		DeepLabv3+    & 73.15 & 84.33 & 79.68 & 88.56 & 54.39 & 70.05 & 69.47 & 81.74 & 50.38 & 66.46 & 65.41 & 78.23 & 83.21\\  
		UNet          & 75.37 & 85.79 & 81.86 & 89.86 & 58.81 & 73.72 & 72.09 & 83.56 & 59.54 & 74.29 & 69.53 & 81.44 & 85.09 \\
		
		PSPNet        & 72.79 & 84.10 & 80.86 & 89.26 & 56.16 & 71.45 & 70.87 & 82.69 & 47.02 & 63.68 & 65.54 & 78.24 & 83.87 \\
		
		\hline
		SANet (ours)   & 75.28 & 85.77 & 82.25 & 90.13 & 58.81 & 73.75 & 72.46 & 83.82 & 53.97 & 69.76 & 68.55 & 80.65 & 85.12\\
		SANet-resnet101 & 77.49 & 87.18 & 84.53 & 91.51 & 61.06 & 75.54 & 73.36 & 84.43 & 60.03 & 74.67 & 71.29 & 82.67 & 86.47\\
		\hline\hline
	\end{tabular}
\end{table*}
\begin{figure*}[!t]
\centering
\subfigure[]{
\begin{minipage}[!t]{0.13\linewidth}
	\centering
	\includegraphics[width=1.1\linewidth]{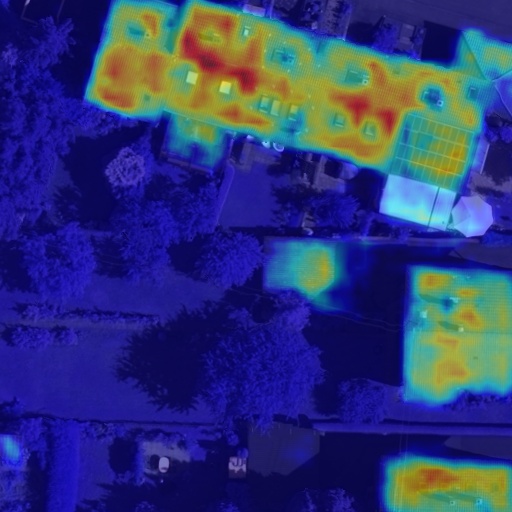}\vspace{4pt}
	\includegraphics[width=1.1\linewidth]{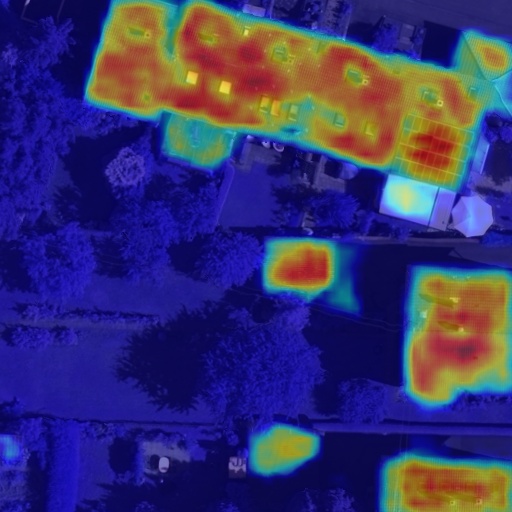}
\end{minipage}	
	}
\subfigure[]{
	\begin{minipage}[!t]{0.13\linewidth}
		\centering
		\includegraphics[width=1.1\linewidth]{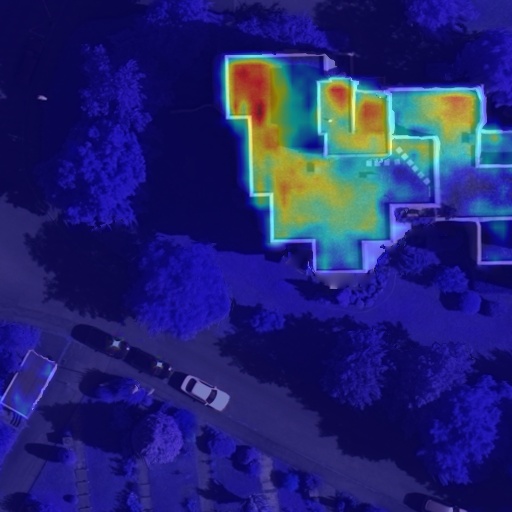}\vspace{4pt}
		\includegraphics[width=1.1\linewidth]{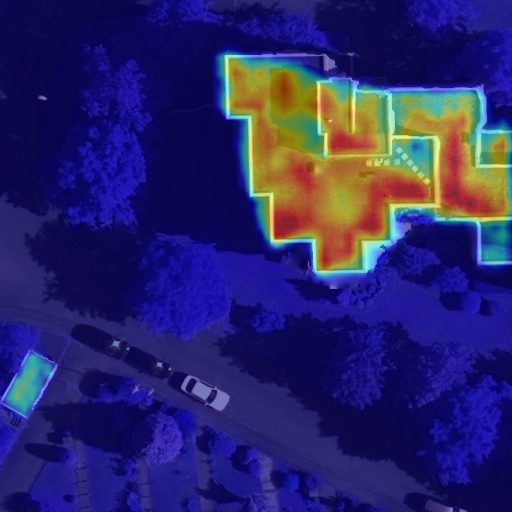}
	\end{minipage}	
}
\subfigure[]{
	\begin{minipage}[!t]{0.13\linewidth}
			\centering
		\includegraphics[width=1.1\linewidth]{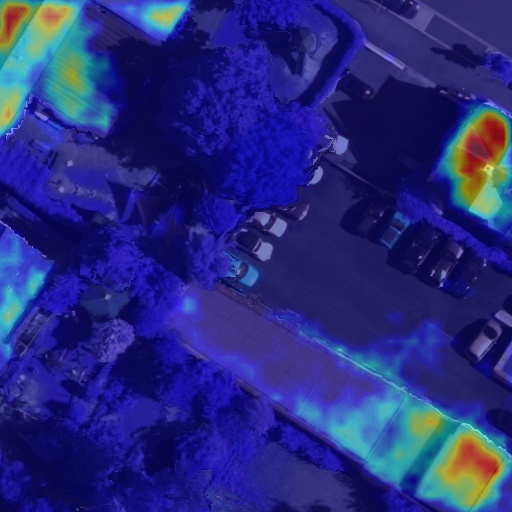}\vspace{4pt}
		\includegraphics[width=1.1\linewidth]{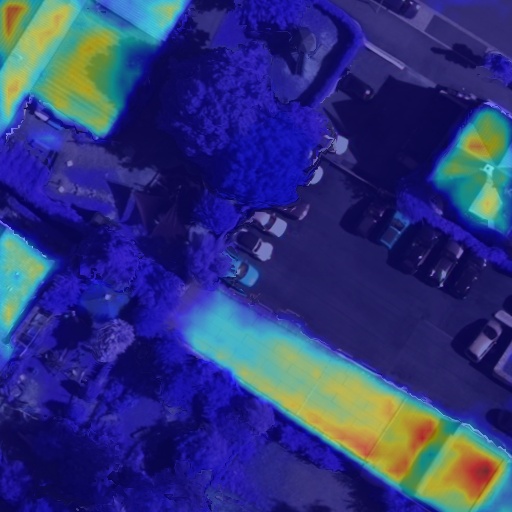}
	\end{minipage}	
}
\subfigure[]{
	\begin{minipage}[!t]{0.13\linewidth}
			\centering
		\includegraphics[width=1.1\linewidth]{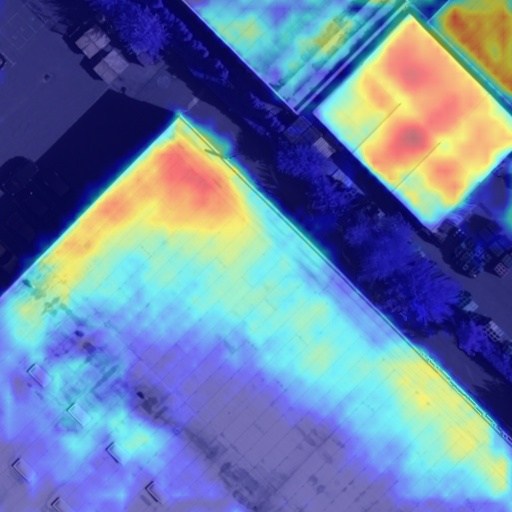}\vspace{4pt}
		\includegraphics[width=1.1\linewidth]{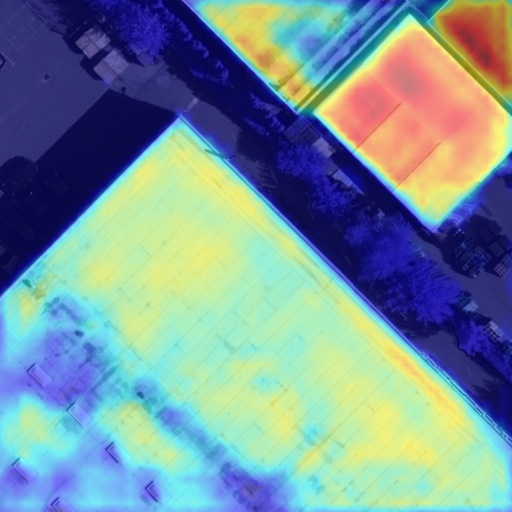}
	\end{minipage}	
}
\subfigure[]{
	\begin{minipage}[!t]{0.13\linewidth}
			\centering
		\includegraphics[width=1.1\linewidth]{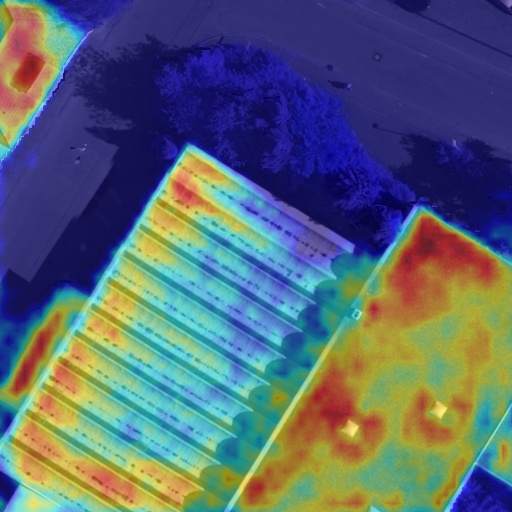}\vspace{4pt}
		\includegraphics[width=1.1\linewidth]{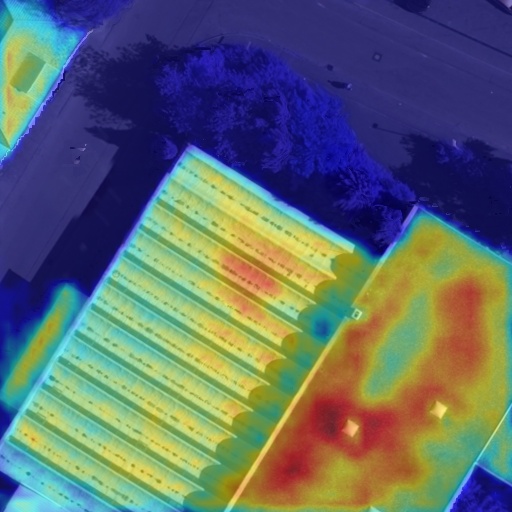}
	\end{minipage}	
}
\subfigure[]{
	\begin{minipage}[!t]{0.13\linewidth}
			\centering
		\includegraphics[width=1.1\linewidth]{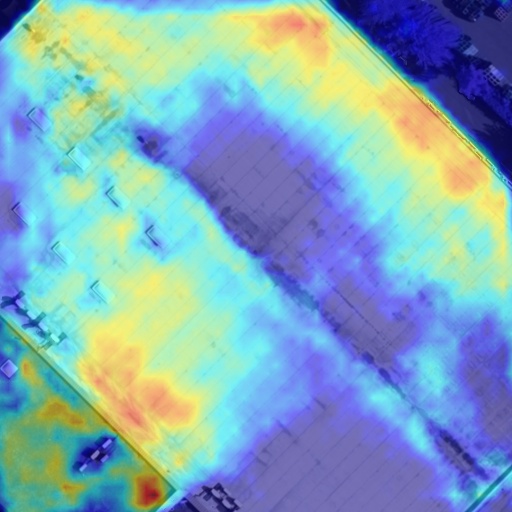}\vspace{4pt}
		\includegraphics[width=1.1\linewidth]{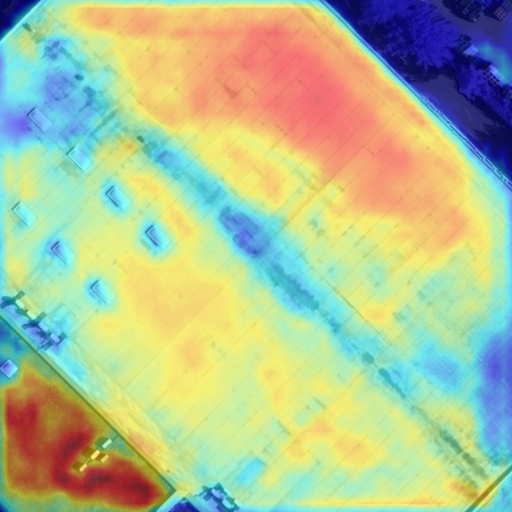}
	\end{minipage}	
}
	\caption{The class activation mappings (CAM) of building with small to large scales, from (a) to (f). The images of the first row is generated by FCN8s and the second row is for FCN8s+SAM-M.}
	\label{fig_cam}
\end{figure*}

\section{Experiments}
\subsection{General Setup}
\subsubsection{Data set Information}
We evaluate the proposed methods on the Vaihinger data set from ISPRS 2D Semantic Labeling Contest, the contest has ended and all the data has been publicly available. The data set contains 33 patches with average size of around $2500 \times 2000$ and the ground sampling distance is 9 $cm$, each consisting of a true orthophoto (TOP) extracted from a larger TOP mosaic. We only use the IRRG images in our experiment without DSM and normalized DSM. The dataset has been classified into six classes, including impervious surfaces, buildings, low vegetation, tree, car and clutter/background. We split the whole dataset into training, validation and testing sets. We follow the test set provided by the official, we adopt five patches (image id 11, 15, 28, 30, 34) as validation set and the rest of patches as training set. We crop the large patches into slices of $512 \times 512$ using 50\% overlapped window.
In terms of the strategy of data argumentation, we only adopt random horizontal and vertical flip. We use per-class Iou, per-class F1-score, mean IoU, mean F1-score, and overall accuracy as our metrics.
\subsubsection{Training Configuration}
All the experiments are implemented with PyTorch 1.1.0, CUDA 9.0 and CuDNN 7. The networks run 200 epochs on single Tesla V100 GPU, using Adam optimizer with weight decay of 2e-4 and momentum of 0.9. The initial learning rate is 5e-4 and we adopt "ploy" learning rate policy with power of 0.9. We set batch size to 16 to fit our GPU memory. In order to alleviate the unbalanced categories, we adopt cross entropy loss with weight $W_{class}=\frac{1}{log(P_{class}+c)}$ and we set $c$ to 1.12. 
\subsection{Ablation studies}
In this section, we conduct sets of ablation studies based on the popular framework FCN8s to demonstrate the effectiveness of the proposed module, and we choose pre-trained ResNet34 as our network backbone. 

We compare both IoU and F1-Score of each network, and the comparison results are shown in Table \ref{table_ablations}.
To valid the effectiveness of SAM, we firstly embed single SAM before the last score layer of FCN8s as FCN8s-SAM-S, and it has slightly boost for each class compared with vanilla FCN8s. Further, we employ SAM at the end of each stage of FCN8s as FCN8s-SAM-M, the large margin improvement demonstrates employing SAM in multi-scale feature maps can get much better performance.
Since our SAM implicitly introduces spatial attention mechanism by considering the re-sampling map as weighted map (as analyzed in Section II.B), and the attention module have the ability let network pay more attention to "what" and "where", does our scale-aware module really work in the experiments? 
In order to prove our method is different from the spatial attention mechanism and its effectiveness, we replace SAM in FCN8s-SAM into spatial attention module while maintaining the number of parameters unchanged, and we name them FCN8s-SAM-SC, FCN8s-SAM-MC for single and multiple version, respectively.
The experimental results demonstrate that the performance of counterpart networks with spatial attention module is slightly better than the vanilla one, but the recognition accuracy of our SAM is much better than the attention module.
Therefore, our approach is much more effective compared to the spatial attention mechanism when faced with objects in various sizes.




\subsection{Visualization Analysis}
To further illustrate the effectiveness of our SAM and elaborate the improvement is not due to the additional parameters, we generate class activation mappings of buildings with different scales for FCN8s and FCN8s-SAM-M.

As the Fig. \ref{fig_cam} shown, the first row is cam for FCN8s and the second row is for FCN8s-SAM-M.
In the first two columns, vanilla FCN8s cannot recognize the small buildings because the feature is dominated by the surrounding objects, but our SAM can not only identify the small buildings but is more sensitive to the areas belong to building than FCN8s. 
The recognition of long strip objects is a challenging task in semantic segmentation, these objects need large receptive field to get enough context information, but the large receptive field with square shape includes too much irrelevant information that usually leads to unexpected predictions.
In the third column, the conventional FCN8s cannot classify the building with long strip shape completely, and our SAM has better response to the entire building.
For large buildings in the forth and fifth columns, the activation response of FCN8s is concentrated at a part of the buildings, but the distribution of SAM is relatively more uniform. Thus, our SAM can perform better on long strip and large objects.
Considering the extreme case, the buildings occupy most of imagery, the lack of receptive field will lead to misclassification. For the extremely large buildings in the last column, 
FCN8s has low response in the central part but high response at the boundaries of buildings, so FCN8s wrongly classifies the central part as imperious surface. However, our SAM also have ideal activation response for the whole building.

\subsection{Evaluation and Comparisons}
As shown in Table \ref{table_evalation}, we compare our SANet with other state-of-the-art networks based on pre-trained ResNet34 (if without  additional notification), including light-weight RefineNet, DeepLabv3, DeepLabv3+, UNet, and PSPNet.
All these networks are designed to get better recognition for objects with various sizes by fusing multi-scale information.
Light-weight RefineNet get worse accuracy compared to other networks. By employing atrous spatial pyramid pooling module, DeepLabv3, DeepLabv3+ and PSPNet get similar accuracy.
Although the performance of UNet is much better than the others, the mirror-like encoder-decoder structure makes it consumes large scale of computational resources.
As the last two rows in Table \ref{table_evalation} shown, our SANet get slightly better accuracy compared to the heavy-weight UNet, and it is worth noting that we only adopt IRRG images as training data without DSM or normalized DSM, these additional features, deeper and wider networks, and the refinement of decoder will further improve the segmentation accuracy of our method.

\section{Conclusion}
In this letter, we proposed a novel scale-aware module (SAM) for semantic segmentation of high-resolution aerial images.
SAM performs re-sampling operation for each pixels based on the combination of re-sampling information and re-sampling map. The re-sampling information is learned from training data, and the re-sampling map is adaptive to the input data. Thus, the networks with SAM not only have a stronger ability to recognize objects with various sizes but also adaptive to the input. Furthermore, our proposed module can be trained end-to-end and easily embedded in the existing network.
We evaluate our method on ISPRS Vaihinger Dataset, and we embed SAM in a simple framework FCN8s.
The experiment and visualization results illustrate that SAM have better recognition performance of objects with inconsistent scales.
Although our scale-aware method get much better performance for the objects with various sizes in remote sensing images, the networks still need reliable context information to classify the adjacent objects in high similarity. In future, we will integrate context-aware ability to the our works that makes the DCNNs have better performance for semantic segmentation of high-resolution remote sensing images.


%

\ifCLASSOPTIONcaptionsoff
  \newpage
\fi



\bibliographystyle{IEEEtran}
\bibliography{IEEEabrv, references}

\begin{thebibliography}{10}
\providecommand{\url}[1]{#1}
\csname url@samestyle\endcsname
\providecommand{\newblock}{\relax}
\providecommand{\bibinfo}[2]{#2}
\providecommand{\BIBentrySTDinterwordspacing}{\spaceskip=0pt\relax}
\providecommand{\BIBentryALTinterwordstretchfactor}{4}
\providecommand{\BIBentryALTinterwordspacing}{\spaceskip=\fontdimen2\font plus
\BIBentryALTinterwordstretchfactor\fontdimen3\font minus
  \fontdimen4\font\relax}
\providecommand{\BIBforeignlanguage}[2]{{%
\expandafter\ifx\csname l@#1\endcsname\relax
\typeout{** WARNING: IEEEtran.bst: No hyphenation pattern has been}%
\typeout{** loaded for the language `#1'. Using the pattern for}%
\typeout{** the default language instead.}%
\else
\language=\csname l@#1\endcsname
\fi
#2}}
\providecommand{\BIBdecl}{\relax}
\BIBdecl

\bibitem{Krizhevsky2012}
A.~Krizhevsky, I.~Sutskever, and G.~E. Hinton, ``{ImageNet} {Classification}
  with {Deep} {Convolutional} {Neural} {Networks},'' in \emph{Advances in
  {Neural} {Information} {Processing} {Systems} 25}, 2012, pp. 1097--1105.

\bibitem{lin_microsoft_2014}
T.-Y. Lin, M.~Maire, S.~Belongie, L.~Bourdev, R.~Girshick, J.~Hays, P.~Perona,
  D.~Ramanan, C.~L. Zitnick, and P.~Dollár, ``Microsoft {COCO}: {Common}
  {Objects} in {Context},'' \emph{arXiv:1405.0312 [cs]}, May 2014.

\bibitem{everingham_pascal_2010}
M.~Everingham, L.~Van~Gool, C.~K.~I. Williams, J.~Winn, and A.~Zisserman,
  ``\BIBforeignlanguage{en}{The {Pascal} {Visual} {Object} {Classes} ({VOC})
  {Challenge}},'' \emph{\BIBforeignlanguage{en}{International Journal of
  Computer Vision}}, vol.~88, no.~2, pp. 303--338, Jun. 2010.

\bibitem{long_fully_nodate}
J.~Long, E.~Shelhamer, and T.~Darrell, ``\BIBforeignlanguage{en}{Fully
  {Convolutional} {Networks} for {Semantic} {Segmentation}},''
  \emph{\BIBforeignlanguage{en}{Proc. Comput. Vis. Pattern Recognit.}}, p.~10,
  Jun. 2015.

\bibitem{ronneberger_u-net:_2015}
O.~Ronneberger, P.~Fischer, and T.~Brox, ``U-{Net}: {Convolutional} {Networks}
  for {Biomedical} {Image} {Segmentation},'' \emph{arXiv:1505.04597 [cs]}, May
  2015.

\bibitem{badrinarayanan_segnet:_2015}
V.~Badrinarayanan, A.~Kendall, and R.~Cipolla, ``{SegNet}: {A} {Deep}
  {Convolutional} {Encoder}-{Decoder} {Architecture} for {Image}
  {Segmentation},'' \emph{arXiv:1511.00561 [cs]}, Nov. 2015.

\bibitem{noh_learning_2015}
H.~Noh, S.~Hong, and B.~Han, ``Learning {Deconvolution} {Network} for
  {Semantic} {Segmentation},'' \emph{arXiv:1505.04366 [cs]}, May 2015.

\bibitem{chen_deeplab:_2016}
L.-C. Chen, G.~Papandreou, I.~Kokkinos, K.~Murphy, and A.~L. Yuille,
  ``{DeepLab}: {Semantic} {Image} {Segmentation} with {Deep} {Convolutional}
  {Nets}, {Atrous} {Convolution}, and {Fully} {Connected} {CRFs},''
  \emph{arXiv:1606.00915 [cs]}, Jun. 2016.

\bibitem{chen_encoder-decoder_2018}
L.-C. Chen, Y.~Zhu, G.~Papandreou, F.~Schroff, and H.~Adam, ``Encoder-{Decoder}
  with {Atrous} {Separable} {Convolution} for {Semantic} {Image}
  {Segmentation},'' \emph{arXiv:1802.02611 [cs]}, Feb. 2018.

\bibitem{lin_refinenet:_2016}
G.~Lin, A.~Milan, C.~Shen, and I.~Reid, ``{RefineNet}: {Multi}-{Path}
  {Refinement} {Networks} for {High}-{Resolution} {Semantic} {Segmentation},''
  \emph{arXiv:1611.06612 [cs]}, Nov. 2016.

\bibitem{peng_large_2017}
C.~Peng, X.~Zhang, G.~Yu, G.~Luo, and J.~Sun, ``Large {Kernel} {Matters} --
  {Improve} {Semantic} {Segmentation} by {Global} {Convolutional} {Network},''
  \emph{arXiv:1703.02719 [cs]}, Mar. 2017.

\bibitem{nogueira_dynamic_2019}
K.~Nogueira, M.~D. Mura, J.~Chanussot, W.~R. Schwartz, and J.~A.~d. Santos,
  ``Dynamic {Multicontext} {Segmentation} of {Remote} {Sensing} {Images}
  {Based} on {Convolutional} {Networks},'' \emph{IEEE Transactions on
  Geoscience and Remote Sensing}, pp. 1--18, 2019.

\bibitem{li_adaptive_2019}
G.~Li, L.~Li, H.~Zhu, X.~Liu, and L.~Jiao, ``Adaptive {Multiscale} {Deep}
  {Fusion} {Residual} {Network} for {Remote} {Sensing} {Image}
  {Classification},'' \emph{IEEE Transactions on Geoscience and Remote
  Sensing}, pp. 1--16, 2019.

\bibitem{lu_feature_2019}
X.~Lu, H.~Sun, and X.~Zheng, ``A {Feature} {Aggregation} {Convolutional}
  {Neural} {Network} for {Remote} {Sensing} {Scene} {Classification},''
  \emph{IEEE Transactions on Geoscience and Remote Sensing}, pp. 1--13, 2019.

\bibitem{jaderberg_spatial_2015}
M.~Jaderberg, K.~Simonyan, A.~Zisserman, and K.~Kavukcuoglu, ``Spatial
  {Transformer} {Networks},'' \emph{arXiv:1506.02025 [cs]}, Jun. 2015.

\bibitem{wei_learning_2017}
Z.~Wei, Y.~Sun, J.~Wang, H.~Lai, and S.~Liu, ``Learning {Adaptive} {Receptive}
  {Fields} for {Deep} {Image} {Parsing} {Network},'' in \emph{2017 {IEEE}
  {Conference} on {Computer} {Vision} and {Pattern} {Recognition} ({CVPR})},
  Jul. 2017, pp. 3947--3955.

\bibitem{zhang_scale-adaptive_2017}
R.~Zhang, S.~Tang, Y.~Zhang, J.~Li, and S.~Yan, ``Scale-{Adaptive}
  {Convolutions} for {Scene} {Parsing},'' in \emph{2017 {IEEE} {International}
  {Conference} on {Computer} {Vision} ({ICCV})}, Oct. 2017, pp. 2050--2058.

\end{thebibliography}
\end{document}